\documentclass[10pt,twocolumn,letterpaper]{article}

\usepackage{cvpr}              

\usepackage{graphicx}
\usepackage{amsmath}
\usepackage{amssymb}
\usepackage{booktabs}
\usepackage{color}
\usepackage{bm}
\usepackage{threeparttable}
\usepackage{multirow}
\usepackage[accsupp]{axessibility}

\usepackage[pagebackref,breaklinks,colorlinks]{hyperref}

\usepackage[capitalize]{cleveref}
\crefname{section}{Sec.}{Secs.}
\Crefname{section}{Section}{Sections}
\Crefname{table}{Table}{Tables}
\crefname{table}{Tab.}{Tabs.}

\def\cvprPaperID{4782} 
\def\confName{CVPR}
\def\confYear{2023}

\begin{document}

\title{Visual-Tactile Sensing for In-Hand Object Reconstruction}

\author{Wenqiang Xu\textsuperscript{* 1, 2}, Zhenjun Yu\textsuperscript{* 1}, Han Xue\textsuperscript{1}, Ruolin Ye\textsuperscript{3}, Siqiong Yao\textsuperscript{1}, Cewu Lu\textsuperscript{\S \ 1, 2}\\
\textsuperscript{1}Shanghai Jiao Tong University  \quad 
\textsuperscript{2}Shanghai Qi Zhi institute  \quad
\textsuperscript{3}Cornell University\\
\textsuperscript{1}{\tt\small \{vinjohn, jeffson-yu, xiaoxiaoxh, yaosiqiong, lucewu\}@sjtu.edu.cn}\\
\textsuperscript{3}{\tt\small ry273@cornell.edu}
}

\twocolumn[{%
\renewcommand\twocolumn[1][]{#1}%
\maketitle
\begin{center}
    \centering
    \captionsetup{type=figure}
    \includegraphics[width=1\linewidth]{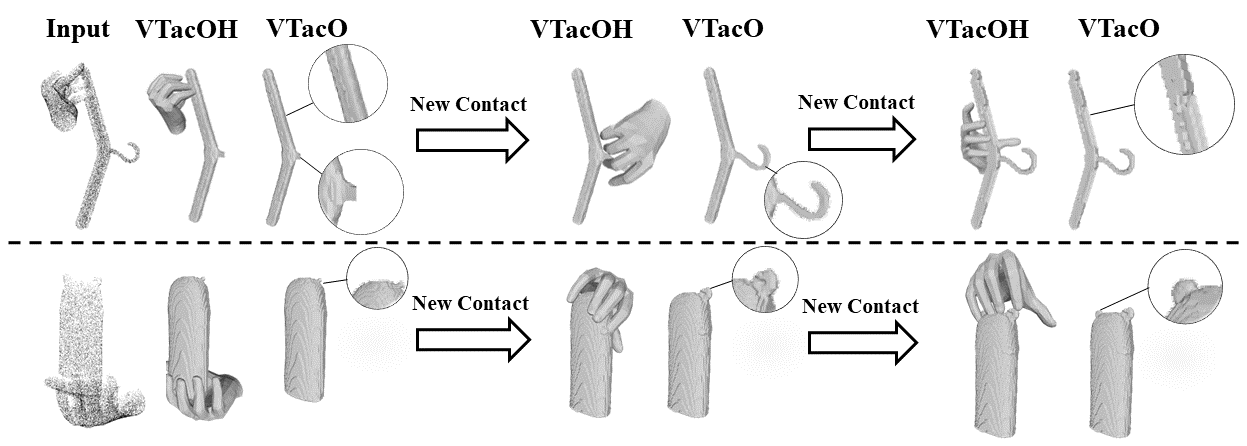}
    \captionof{figure}{Our proposed visual-tactile learning framework \textbf{VTacO} and its extended version \textbf{VTacOH} can reconstruct both the rigid and non-rigid in-hand objects. It also supports refining the mesh in an incremental manner.}
    \label{fig:teaser}
\end{center}%
}]

\begin{abstract}
    Tactile sensing is one of the modalities humans rely on heavily to perceive the world. Working with vision, this modality refines local geometry structure, measures deformation at the contact area, and indicates the hand-object contact state.  
   With the availability of open-source tactile sensors such as DIGIT, research on visual-tactile learning is becoming more accessible and reproducible. 
   Leveraging this tactile sensor, we propose a novel visual-tactile in-hand object reconstruction framework \textbf{VTacO}, and extend it to \textbf{VTacOH} for hand-object reconstruction. Since our method can support both rigid and deformable object reconstruction, no existing benchmarks are proper for the goal. We propose a simulation environment, VT-Sim, which supports generating hand-object interaction for both rigid and deformable objects. With VT-Sim, we generate a large-scale training dataset and evaluate our method on it. Extensive experiments demonstrate that our proposed method can outperform the previous baseline methods qualitatively and quantitatively. Finally, we directly apply our model trained in simulation to various real-world test cases, which display qualitative results.
   Codes, models, simulation environment, and datasets are available at \url{https://sites.google.com/view/vtaco/}.
\end{abstract}

\renewcommand{\thefootnote}{}
\footnotetext{* indicates equal contributions. \\ \indent \indent \S \ Cewu Lu is the corresponding author, the member of Qing Yuan Research Institute and MoE Key Lab of Artificial Intelligence, AI Institute, Shanghai Jiao Tong University, China and Shanghai Qi Zhi institute.}

\section{Introduction}

Human beings have a sense of object geometry by seeing and touching,
especially when the object is in manipulation and undergoes a large portion of occlusion, where visual information is not enough for the details of object geometry. In such cases, vision-based tactile sensing is a good supplement as a way of proximal perception. 
In the past, few vision-based tactile sensors were commercially available or open-source, so the visual-tactile sensing techniques could not be widely studied. 
Previous works \cite{inhand1,interactionfusion} on in-hand object reconstruction either studied rigid objects or were limited to simple objects with simple deformation. 


Vision-based tactile sensors \cite{digit,gelsight,gelslim,gelsight_wedge} can produce colorful tactile images indicating local geometry and deformation in the contact areas. 
In this work, we mainly work with DIGIT \cite{digit} as it is open-source for manufacture and is easier to reproduce sensing modality.
With tactile images, we propose a novel \textbf{V}isual-\textbf{Tac}tile in-hand \textbf{O}bject reconstruction framework named \textbf{VTacO}. VTacO reconstructs the object geometry with the input of a partial point cloud observation and several tactile images. The tactile and object features are extracted by neural networks and fused in the Winding Number Field (WNF) \cite{wnf}, and the object shape is extracted by Marching Cubes algorithm \cite{marching_cubes}. WNF can represent the object shape with open and thin structures. The poses of tactile sensors can be determined either by markers attached or by hand kinematics. By default, VTacO assumes the tactile sensor poses can be obtained independently, but we also discuss how to obtain the tactile sensor poses alongside the object with hand pose estimation. The corresponding method is named \textbf{VTacOH}. 

With tactile information, we can enhance pure visual information from three aspects: (1) \textbf{Local geometry refinement}. We use tactile sensing as proximal perception to complement details of local geometry. (2) \textbf{Deformation at contact area}. Objects, even those we consider rigid, can undergo considerable deformation given external forces exerted by hand. (3) \textbf{Hand-object contact state}. Tactile sensors indicate whether the hand is in contact with the object's surface. To demonstrate such merits, we conduct the object reconstruction tasks in both rigid and non-rigid settings.

Since obtaining the ground truth of object deformation in the real world is hard, we first synthesize the training data from a simulator. DIGIT has an official simulation implementation, TACTO \cite{tacto}. However, it is based on pybullet \cite{pybullet}, which has limited ability to simulate deformable objects. Thus, we implement a tactile simulation environment \textbf{VT-Sim} in Unity. In VT-Sim, we generate hand poses with GraspIt! \cite{graspit}, and simulate the deformation around the contact area with an XPBD-based method. In the simulation, we can easily obtain depth image, tactile image, DIGIT pose, and object WNF as training samples for both rigid and non-rigid objects.

To evaluate the method, we compare the proposed visual-tactile models with its visual-only setting, and the previous baseline 3D Shape Reconstruction from Vision and Touch (3DVT) \cite{3dvt}. Extensive experiments show that our method can achieve both quantitative and qualitative improvements on baseline methods.
Besides, since we make the tactile features fused with winding number prediction, we can procedurally gain finer geometry reconstruction results by incrementally contacting different areas of objects. 
It can be useful for robotics applications \cite{robot1,robot2}.
Then, we directly apply the model trained with synthesis data to the real world. It shows great generalization ability.

We summarize our contributions as follows:
\begin{itemize}
    \item A visual-tactile learning framework to reconstruct an object when it is being manipulated. We provide the object-only version VTacO, and the hand-object version VTacOH.
    \item A simulation environment, VT-Sim, which can generate training samples. We also validate the generalization ability of the models trained on the simulated data to the real-world data.
\end{itemize}

\section{Related Works}
\paragraph{Vision-based Tactile Sensors}
Tactile sensors have been studied by robotics and material communities for decades. As a result, different approaches to mimic human tactile sensing are developed. Among them, the most relevant approaches are the vision-based tactile sensors \cite{gelsight,gelslim,gelsight_wedge,digit}. These methods typically use a gel to contact the object, and a camera sensor to observe the deformation through the layer. 
If the gel is calibrated, the deformation corresponds with a force so that we can obtain the contact force by observing the tactile images. Among these sensors, DIGIT~\cite{digit} is the one with the structure most suitable for the human hand. The fact that it is attached to fingers rather than wrapping around makes it aligned with human fingertip structures. 

Vision-based tactile sensors, by nature, have the merits of high-density, multi-DOF force sensing and a data format of image, which is suitable for neural network processing. These features make them beneficial for the computer vision community. However, it could be hard to re-implement most tactile sensors without open-source circuit design and drive programs. 
These essential components are necessary for the development of visual-tactile research. As an open-source hardware platform, DIGIT \cite{digit} takes the first step toward addressing this bottleneck problem.



\paragraph{Visual-Tactile Sensing for Object Reconstruction} 
Visual information usually provides global and coarse shape information in object reconstruction with visual-tactile sensing, while tactile information refines local geometry. ~\cite{vtrecon1,shapemap,3dvt}. 

Among these works, ~\cite{3dvt} is most relevant to our method. It jointly processes visual and tactile images to estimate global and local shapes. However, it is limited to watertight objects due to its SDF-based shape modeling approach. Besides this limitation, previous works \cite{vtrecon1,shapemap,robot2} are also limited to static object settings. To our knowledge, none of the previous works considers hand-object interaction and thus ignores the deformation during manipulation.


\paragraph{In-hand Object Reconstruction} 
Earlier works usually consider interacting with rigid \cite{rigid1} or articulated objects \cite{articulate1}. Most of them require prior knowledge about the objects. \cite{inhand1} studied jointly tracking a deformable object and hand during interaction, with required knowledge about object model and texture.

With the development of deep learning recently, the learning algorithms~\cite{obman,grab,ganhand,honnotate} now have the potential to generalize to unseen objects. Among them, InteractionFusion \cite{interactionfusion} which also studies the deformation during manipulation is the most relevant one to our method. However, it adds a requirement for tracking the object from its rest state to model contact deformation.


\section{Method}
\subsection{Overview}
We consider the problem of 3D reconstruction of in-hand objects from a visual-tactile perspective, with which we propose \textbf{VTacO}, a novel deep learning-based visual-tactile learning framework.

In this setting, we assume the visual input is a 3D point cloud $\mathcal{P}$ of an in-hand object converted from a depth image. The tactile input $\mathcal{T}=\{T_i\}^N_{i=1}$ contains $N$ tuples of tactile readings obtained from the sensors, where $T_i=\{(I_i, \bm{p}_i)\}$, $I_i$ is the RGB tactile image, and $\bm{p}_i$ is the 6DOF sensor pose.

The object shape is modeled by the winding number field (WNF), a variant of the signed distance field (SDF). Thus, we can also learn the object representation similar to the SDF-based pipelines \cite{convocc}. The point cloud $\mathcal{P}\in \mathbb{R}^{N_p\times 3}$ is converted to a global feature $\mathcal{F}_p\in \mathbb{R}^{d_p}$ through a feature extractor $\bm{f}_p(\cdot)$, where $N_p$ is point number and $d_p$ is the feature dimension. The tactile image $I_i$ is forwarded to another feature extractor $\bm{f}_t(\cdot)$ to obtain local feature $\mathcal{F}_t \in \mathbb{R}^{d_t}$, where $d_t$ is the feature dimension. Then, we sample $M$ positions $\mathcal{X}=\{\bm{X}_j=(x_j,y_j,z_j)\}_{j=1}^M$ in the 3D space, and fuse $\mathcal{F}_p$ to each position, $\mathcal{F}_t$ to the near positions according to the sensor pose $\bm{p}$. The final feature will be passed to a point-wise decoder to predict the winding number value, which will be used for extracting the shape by Marching Cubes algorithm. The details will be discussed in Sec. \ref{sec:vtaco}

Considering the DIGIT sensor does not provide the sensor pose $\bm{p}$, we provide two different options to obtain $\bm{p}$ in the real world, namely by attaching markers or inferring from hand kinematics. For the latter option, we propose a pipeline \textbf{VTacOH} to jointly estimate the object shape, hand, and sensor pose. It will be discussed in Sec. \ref{sec:vtacoh}.

\begin{figure*}[t!]
    \centering
    \includegraphics[width=1\linewidth]{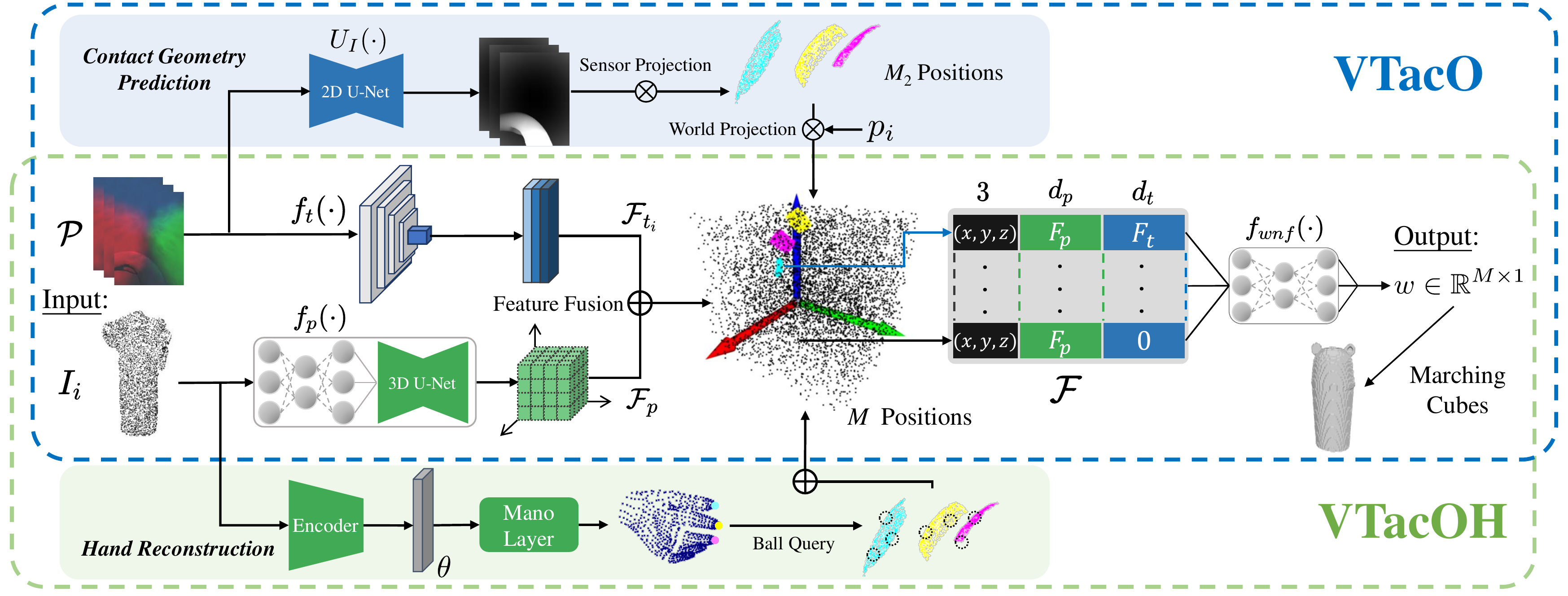}
    
    \caption{\textbf{VTacO} and \textbf{VTacOH}. The input is a set of point clouds and tactile images obtained by tactile sensors. We first compute the global point cloud information with a volume encoder and generate a local feature with a tactile feature encoder. For the position sampling, in \textbf{VTacO}, we predict the local point cloud with a U-Net from the tactile signals and project them to the visual camera space. And in \textbf{VTacOH} we reconstruct the hand by encoding MANO parameters and query positions around the local point clouds generated from the sensors on hand tips. For $M$ sampled positions we first fuse the global and local features, then we estimate the winding number with the WNF decoder of several MLPs. For object reconstruction, we conduct marching cubes algorithm.}
    \label{fig:pipeline}
\end{figure*}

An overview of our approach is illustrated in Fig \ref{fig:pipeline}. 

\begin{figure}[h!]
    \centering
    \includegraphics[width=0.8\linewidth]{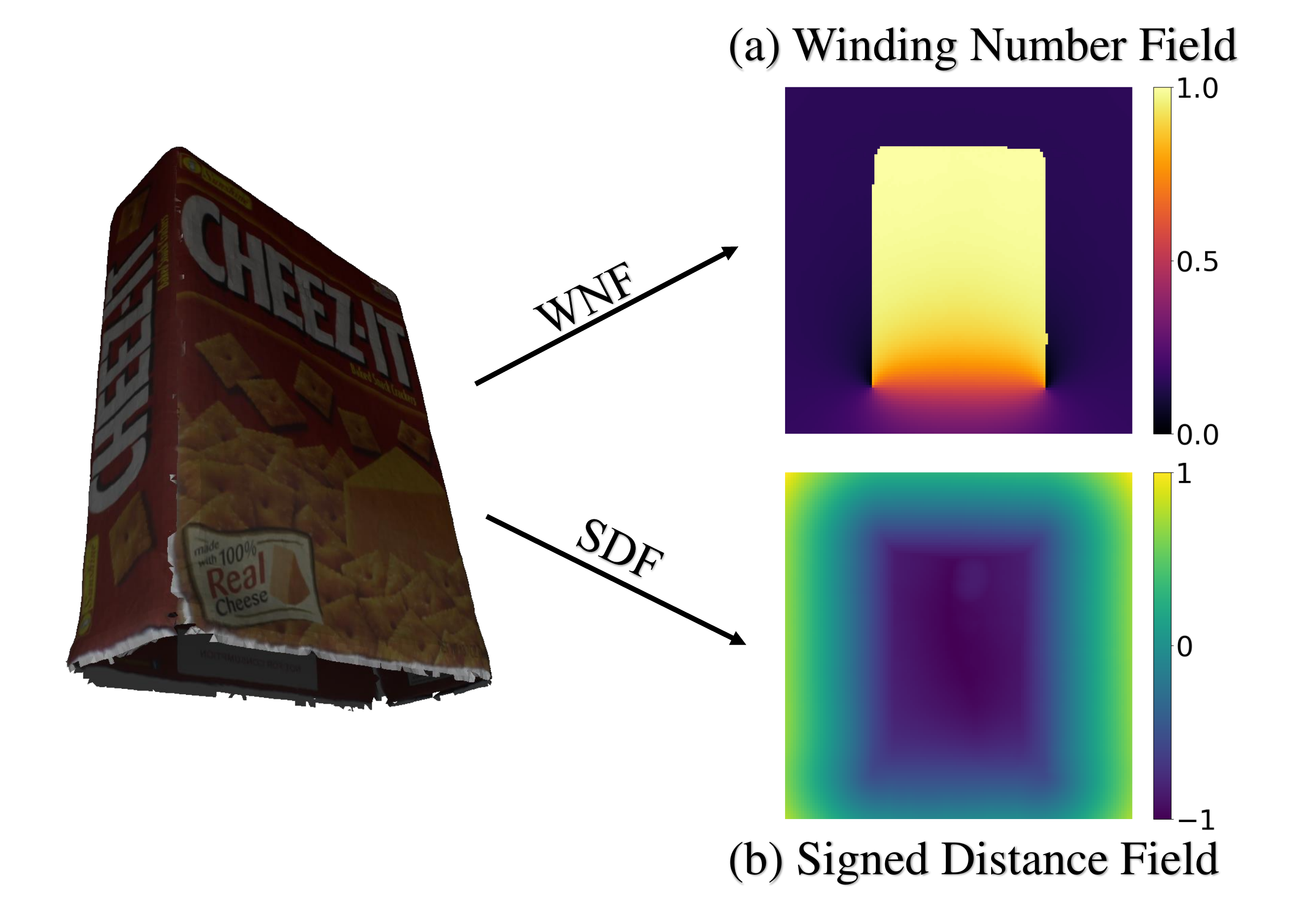}
    \caption{Object in \textbf{WNF} and \textbf{SDF} representation. For an object, the on-body surface is near 1, the off-body area is near 0 and the opening region is around 0.5. While SDF considers only the inside ($<0$) and outside ($>0$).}
    \label{fig:wnf}
\end{figure}

\subsection{Visual-Tactile Object Reconstruction, VTacO}\label{sec:vtaco}

\textbf{Winding Number Field, WNF}
We use WNF to model the object, it is proposed by \cite{wnf} which can implicitly represent a certain surface of an object. 
Unlike how SDF models space occupancy, WNF tracks the solid angles of the surface. Generally, given a Lipschitz surface $S$ and a reference point $p \in \mathbb{R}^3$, if $p$ is inside $S$, the winding number equals 1, and 0 if $p$ outside. The winding number can also be used to model open, non-manifold surfaces and thus can model fine local geometry like thin structures, holes \etc. More details of WNF can be referred to \cite{wnf}. The difference between WNF and SDF is illustrated in Fig. \ref{fig:wnf}.
However, the explicit surface can also be extracted by Marching Cubes algorithm from WNF.

\textbf{Visual Feature Encoder.} Given the input point cloud $\mathcal{P}$, we adopt a 3D UNet-based volume encoder as the feature extractor $\bm{f}_p(\cdot)$. After six fully connected layers, the point cloud features go through a 3D-UNet \cite{3dunet} with four layers of both downsampling and upsampling, and we acquire the global feature $\mathcal{F}_p \in \mathbb{R}^{d_p}$ for every position in the space. We discuss different visual feature encoder choices in Sec. \ref{sec:ablative}.


\textbf{Tactile Feature Encoder.}
We extract the feature $\mathcal{F}_{t_i} \in \mathbb{R}^{d_t}$ from the tactile images $I_i$ with $\bm{f}_t(\cdot)$. Though we can use different backbones for $\bm{f}_t(\cdot)$, in practice, we adopt ResNet-18 \cite{resnet}.

\textbf{Position Sampling \& Feature Fusion.}
Then, we sample $M$ positions in 3D space to query the WNF state later through the decoder. We first try to find the positions relevant to tactile observations.

For each sensor $i$, we assume its sensor pose $\bm{p}_{i}$ is known individually. As we are aware that the tactile sensor like DIGIT itself cannot measure its pose in the real world, we will discuss the solution later (Sec. \ref{sec:sensor_pose}). Then, we estimate the depth of every pixel in each sensor's observation $I_i$ using a simple 2D UNet $U_I(\cdot)$ with 3-layer CNN as an encoder. The relationship between the tactile pixel and the depth is associated with the gel deformation, which can be considered as a simple linear mapping. The physics behind it will be discussed in supplementary materials.
Thus, the estimated depth represents contact geometry. We transform it from tactile sensor space to visual sensor space. In this way, we have $M_1$ points in 3D space which are relevant to tactile images.

Then, we uniformly sample $M_2$ points in the 3D space, together with the $M_1$ positions, giving us $M=M_1 + M_2$ positions to query.

For each position in $M_1$, we fuse the location $X_j$ with visual feature $\mathcal{F}_p$, and tactile feature $\mathcal{F}_t$. And in $M_2$, we substitute the tactile feature with all-0 vectors. We discuss different fusion strategies in Sec. \ref{sec:ablative}.  After fusing, we have position-wise feature $\mathcal{F} \in \mathbb{R}^d$, where $d$ is the dimension of the fused features.
 


\textbf{WNF Decoder.} Finally, we use a 5-layer MLP $f_{wnf}(\cdot)$ as a decoder to predict the winding number $w$ for each query, with $w = f_{wnf}(\mathcal{F})$.

\textbf{Loss Function} We train the whole pipeline in two parts. 

First, we train $U_I$ with pixel-wise $L_1$ loss. We can obtain the ground truth of the depth images for tactile sensors in simulation, but cannot obtain them in the real world. Thus, when dealing with real-world images, we directly apply the model learned in the simulation.

Then, we jointly train $\bm{f}_p$, $\bm{f}_t$, $\bm{f}_{wnf}$ with $L_1$ loss to supervise the winding number.
\begin{equation} \label{eq:l_wnf}
    \mathcal{L}_w = |w - w^*|.
\end{equation}

\subsection{Sensor Pose Estimation}\label{sec:sensor_pose}
As aforementioned, the DIGIT sensor cannot measure its pose in the real world. To address this, we provide two options: attaching markers and inferring from hand kinematics.

\subsubsection{Estimating Sensor Pose By Markers}
Following the practice in pose estimation, we print ArUco tags \cite{aruco} and attach them to fingertips. Then, we estimate the fingertip poses by estimating the pose of tags. We adopt OpenCV \cite{opencv} to implement the pose estimation. It is a common practice, thus we leave the details in the supplementary materials.

\begin{figure}[h!]
    \centering
   \includegraphics[width=0.8\linewidth]{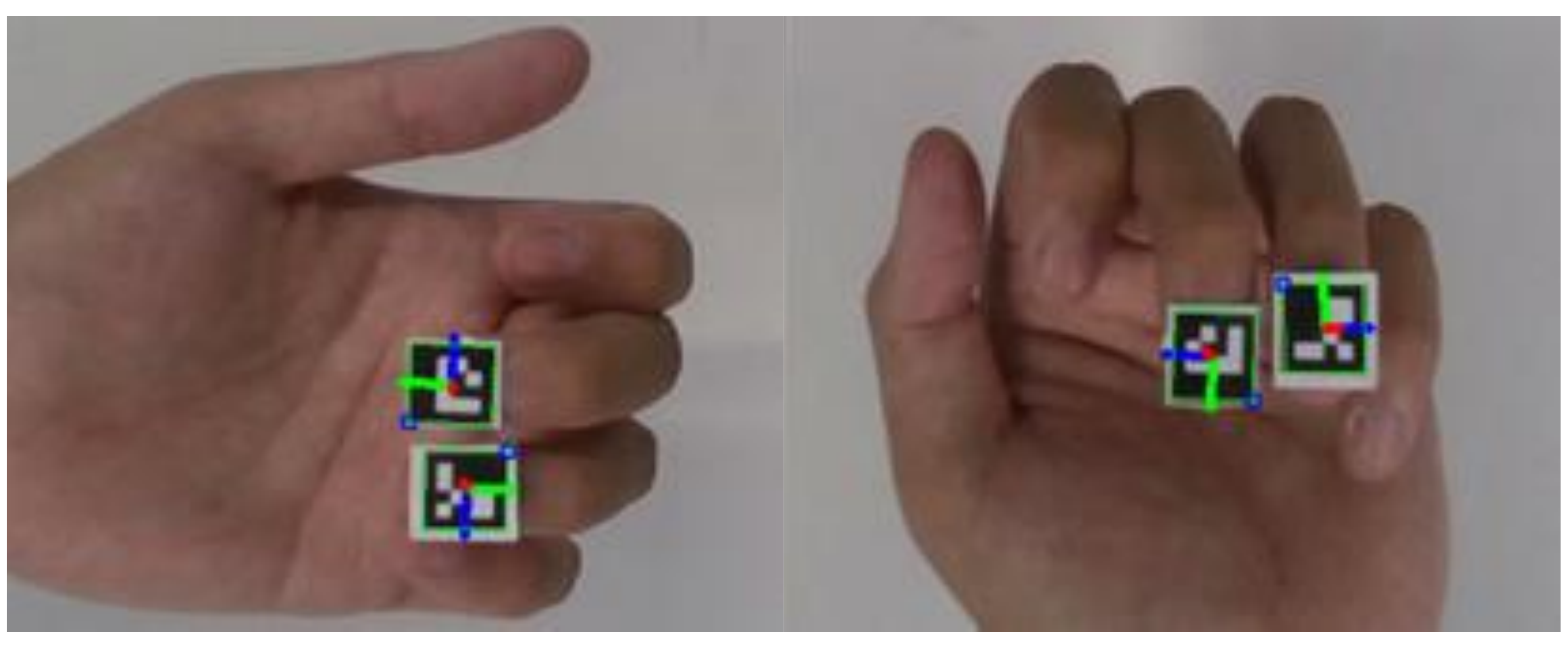}
    \caption{Estimating the fingertips' pose from the attached ArUco markers.}
    \label{fig:sensor_pose}
    \vspace{-0.8cm}
\end{figure}

\subsubsection{Estimating Sensor Pose With Hand Kinematics}\label{sec:vtacoh}
Since the input $\mathcal{P}$ contains a view of the in-hand object, it allows us to reconstruct the hand alongside the object. We can infer the poses of tactile sensors based on the prior that they are attached to the fingertips. We can infer the poses with hand kinematics.
To jointly learn visual-tactile hand-object reconstruction, we introduce \textbf{VTacOH}.

\paragraph{Hand-Sensor Pose Estimation.}
To reconstruct the hand, we adopt the parametric MANO model \cite{mano} which represents the hand shape with $\beta \in \mathbb{R}^{10}$, and the pose with $\theta \in \mathbb{R}^{51}$, which contains the 6DOF wrist joint position and rotation and axis-angle rotation for 15 joints. Based on the MANO model, we regard a tactile sensor as a fixed extension adhered to the tip. The relative pose between the sensors and tips should remain constant. By recording the local transformation between the sensors and the tip positions, we can easily obtain the sensor pose after having the positions of the tips $\hat{J}_h^{tip} \in \mathbb{R}^{5 \times 3}$ through a differentiable MANO layer with $\theta$ and $\beta$.



Suppose the hand tips positions $\hat{\bm{J}}_h^{tip}$ are predicted, then we have the translation $\bm{t}_h \in \mathbb{R}^{5 \times 3}$ from the hand tips to the DIGIT sensor, giving us predicted DIGIT positions $\bm{\hat{p}} = \hat{\bm{J}}_h^{tip} + \bm{t}_h$

\textbf{Tactile Position Sampling with Hand Prediction.}
It is known that hand pose prediction can be inaccurate due to the error in neural network regression, which also influences the accuracy of the tactile sensor poses. Therefore, the tactile position sampling mentioned in Sec. \ref{sec:vtaco} cannot be directly applied. Here, we present an easy but effective way to handle such circumstances. For every inferred DIGIT position $\bm{\hat{p}}_i$,  we simply query the sphere centered in the local point clouds generated from the tactile sensors with a radius $r=0.1$. The feature fusion remains the same.

\textbf{Loss Function} In VTacOH, we will add additional supervision for hand reconstruction. It follows the common practice of MANO hand estimation by constraining the prediction of axis-angle of the hand joints with $L_2$ loss and we obtain:

\begin{equation} \label{eq:l_hand}
    \mathcal{L}_{hand} = ||\theta - \theta^*||_2^2
\end{equation}


\section{VT-Sim}\label{sec:sim}
We adopt a sim-to-real transfer paradigm to learn the visual-tactile learning task since obtaining information about object deformation in the real world is hard. And recently, many simulation environments have proved to be effective on softbody object simulation with Unity \cite{rfuniverse, rcare}.

We present \textbf{VT-Sim}, a simulation environment based on Unity to produce training samples of hand-object interaction with ground truths of WNF, visual depth images, tactile signals, and sensor poses effectively.

The dataset synthesis process is divided into two parts: \textbf{Hand-pose Acquisition} and \textbf{Interaction Simulation}. 


\begin{figure}[h!]
    \centering
    \includegraphics[width=0.8\linewidth]{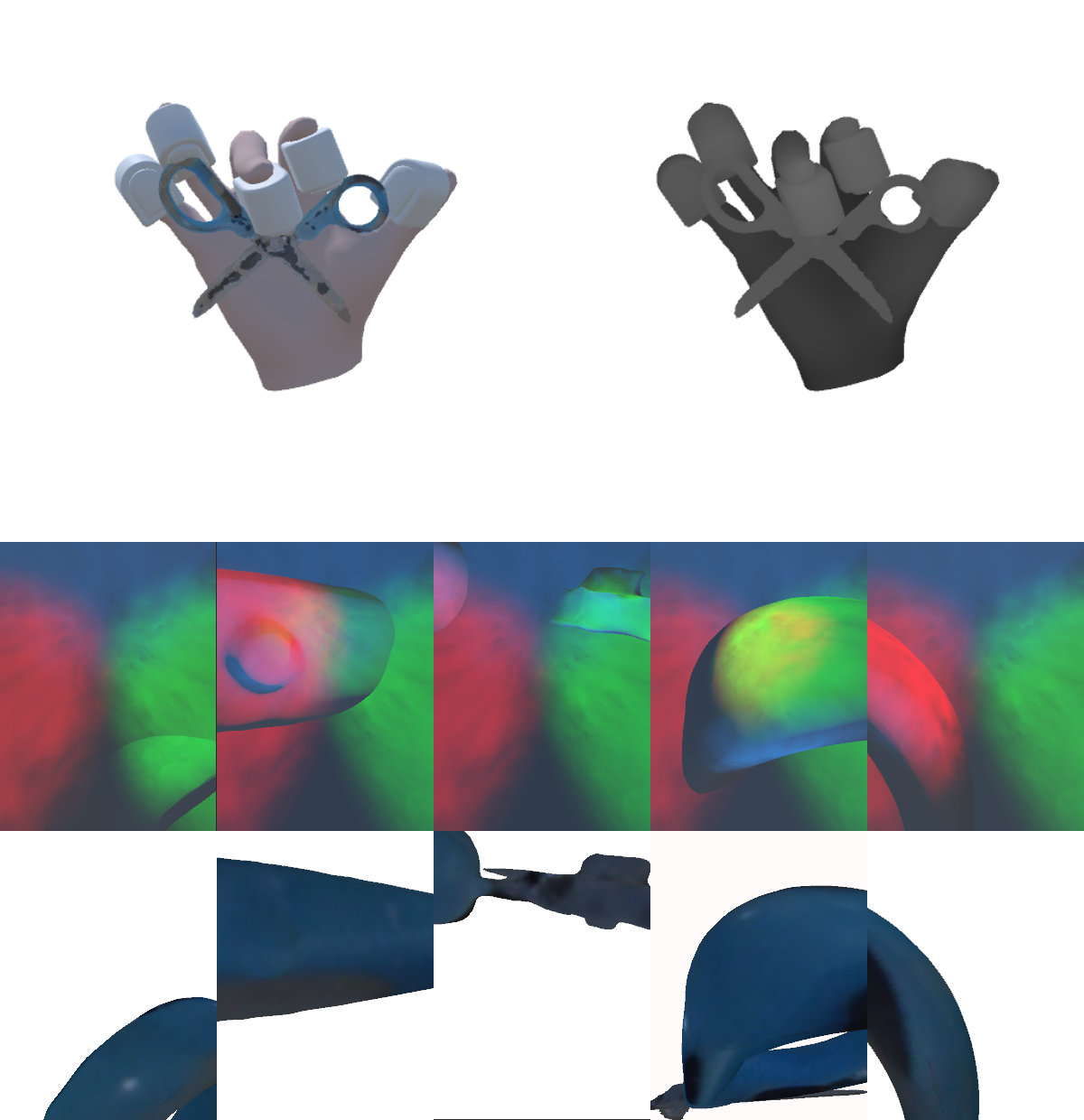}
    \caption{\textbf{VT-sim} example. The DIGIT sensors are attached to the finger tips. When grasping the object, the sensor gel applies forces and collides with the surface mesh, and the sensor camera captures images. The above demonstrates the RGB-D images from a certain view and the difference between tactile images with or without the light in the sensor.}
    \label{fig:vt-sim}
\end{figure}

\textbf{Hand-pose Acquisition} We use GraspIt! \cite{graspit} and a 20-DOF hand model from GraspIt! assets to generate the grasp poses for the selected object models, and save the configuration of hand and objects after the grasp search is over. Since the MANO hand is slightly different from the hand we used to obtain the poses, we use a simple retargeting approach: We calculate the tip positions from the saved configuration of the 20-DOF hand model by forward kinematics. Then we align the tips between the 20-DOF hand and the tips of the MANO hand with sensors and apply inverse kinematics to obtain the poses.

\textbf{Interaction Simulation} After obtaining the poses of MANO hands, sensors, and objects, we replay the configurations in Unity with collision. For the rigid object, we add the Rigidbody property and a mesh collider to it. For the soft body object, we model them with Obi \cite{obi}, an XPBD-based physics engine. When we move the MANO hand with sensors to the retargeted pose, the hand sensor will collide with the object and forms the grasp. If the object is soft around the contact area, the contact region will be deformed.

\textbf{Recording Setup} We transplant the implementation of the DIGIT camera from TACTO \cite{tacto} to obtain tactile images. Besides, we set visual cameras on 8 different viewpoints to generate RGB-D images. The point clouds can be converted from the RGB-D images with camera poses. We utilize the library igl \cite{libigl} to generate the winding number field from the object meshes.


Finally, through one VT-Sim step, we obtain a full data point for training, which includes input point cloud $\mathcal{P}$, tactile signals $\mathcal{T}=\{T_i\}^N_{i=1}$ for every tactile sensor, hand poses $\theta^*$ and winding number $w(x)$ for every $x \in \mathbb{R}^3$.

\section{Experiments}\label{sec:exp}
\subsection{Implementation}\label{sec:impl}
We set $N_p = 3000$ as the input point cloud $\mathcal{P}$ converted from the depth image, and randomly sample 100,000 positions in the space with 20,000 on the surface of the object, and subsample to $M = 2048$ for the WNF prediction. An RGB tactile image and depth obtained in VT-Sim have the size of $300 \times 200$. During the training, we set the dimension of global and local features $d_p = d_t = 32$. We set the resolution of voxelization from the visual feature encoder $D = 64$.

We first train the depth and DIGIT pose prediction part with batch size 12 and $LR = 1e-4$ with Adam optimizer for 100 epochs and use the pre-trained parameters for VTacO training. For the training of VTacO, we set the batch size as 6 and $LR = 2e-4$ with Adam optimizer to train on the whole simulated dataset for 100 epochs, and finetune on every type of object with batch size 4 and $LR = 1e-4$ for 50 to 100 epochs according to different types.  For the VTacOH model, we train the hand reconstruction in 100 epochs with batch size 16 and $LR = 1e-4$ with Adam optimizer, then apply the model parameters on feature embedding and train the whole VTacOH with the same hyperparameters and process as VTacO. The overall time for training VTacO and VTacOH on RTX TITAN XP are around 24 hours.

\subsection{Datasets}\label{sec:dataset}
We use two object model benchmarks which provide full model meshes and texture for our reconstruction task.
\textbf{YCB Objects Models} YCB benchmark \cite{ycb} contains 77 object models, with 600 RGBD images from different angles of cameras and their calibration information. For each example model, it includes textured meshes generated with both Poisson reconstruction and volumetric range image integration, a point cloud of the object generated by all the viewpoints. For our experiments, we select 6 object models in the class \textbf{Box} for generating a simulated dataset.

\textbf{AKB-48 Objects Models} AKB-48 benchmark \cite{akb48} is a large-scale articulated object knowledge base that consists of 2,037 real-world 3D articulated object models of 48 categories. For every object, it provides textured models in both whole and part meshes, and a full point cloud from all viewpoints. In both simulated and real experiments, we select 5 to 10 objects in the following 4 categories: \textbf{Bottle}, \textbf{Foldingrack}, \textbf{Lock}, \textbf{Scissor}.

\textbf{Simulated Dataset} With VT-Sim, we can generate adequate datasets based on two object model benchmarks. We consider the categories \textbf{Bottle} and \textbf{Box} as deformable objects with stiffness 0.2 and 0.3, and others as rigid. For every object, we first generate 100 hand poses from \textbf{Hand-pose Acquisition} in VT-Sim, and select 10 to 40 poses that can generate at least one successful touch from five fingertips' Digits. With 8 different views from the RGB-D camera in Unity, we generate 8 point cloud sets from the depth image for every successful grasp. The overall dataset contains more than 10000 data points from 5 categories of object models, with every category, varying from 1000 to 3000. We split the dataset into 7500 for training, 1500 for testing, and 1000 for validation and visualization.

\subsection{Metrics}\label{sec:metrics}
We adopt several metrics to evaluate our method quantitatively: (1) \underline{Intersection over Union (IoU)} over object volumes. (2) \underline{Chamfer Distance (CD)}, and (3) \underline{Earth Mover’s Distance (EMD)} between predicted vertices of mesh and the ground truth; 

\subsection{Results}
We conduct plenty of experiments to show the capability of VTacO. With the above metrics, we compare our results with the visual-tactile baseline 3D Shape Reconstruction from Vision and Touch (denoted ``3DVT'') \cite{3dvt}. Meanwhile, since ``3DVT'' didn't study the reconstruction of deformable objects, we only compare the 3 non-deformable categories: \textbf{Foldingrack}, \textbf{Lock}, \textbf{Scissor}. We also report the object reconstruction result for VTacOH setting. Due to the accurate hand pose estimation, the results of VTacOH are only slightly worse than the VTacO where the sensor poses are the ground truth. The results are shown in Tab. \ref{tab:metric}. We can see that our method outperforms others in most categories in the metrics. Qualitative results are demonstrated in Fig. \ref{fig:qual}. As illustrated in both quantitative and qualitative results, VTacO has better reconstruction results. We speculate that our outperformance is first due to the point-wise approach we conducted, and WNF being a continuous function concerning positions in the space. More importantly, the Feature fusion method can better demonstrate the contact area and the subtle texture of the surfaces. In Sec. \ref{sec:ablative} we will elaborate more to demonstrate the significant effects after we introduce the tactile sensing, and to prove the effectiveness of our choice of encoder and feature fusion strategy.

\begin{table*}[!ht] 

    \center
    \begin{tabular}{c||c|c|c|c|c|c|c}
        \toprule
        Metrics & Method & bottle & box & foldingrack & lock & scissor & mean \\ \hline
        \multirow{4}{*}{IoU}& 3D Vision and Touch
        &            * &            * &        0.746 &        0.468 &        0.691 &        0.622 \\ 
        \multirow{4}{*}{ }  & Ours (Vision only)
        &        0.884 &        0.934 &\textbf{0.837}&        0.877 &        0.763 &        0.857 \\ 
        \multirow{4}{*}{ }  & Ours (VTacO)
        &\textbf{0.887}&\textbf{0.950}&        0.782 &\textbf{0.916}&\textbf{0.777}&\textbf{0.860} \\
        \multirow{4}{*}{ } & Ours (VTacOH)  &  0.886 & 0.947 & 0.765 & 0.911 & 0.772 & 0.855\\
        \midrule
        
        \multirow{4}{*}{CD}& 3D Vision and Touch
        &            * &            * &\textbf{0.242}&        2.631 &        3.206 &        1.268 \\ 
        \multirow{4}{*}{ } & Ours (Vision only)
        &        1.109 &        0.459 &        1.579 &        1.140 &        7.549 &        1.472 \\ 
        \multirow{4}{*}{ } & Ours (VTacO)
        &\textbf{0.936}&\textbf{0.305}&        1.360 &\textbf{0.932}&\textbf{0.894}&\textbf{0.798} \\ 
        \multirow{4}{*}{ } & Ours (VTacOH) & 0.948 & 0.312 & 1.432 & 0.955 & 0.945 & 0.916\\
        \midrule
        
        \multirow{4}{*}{EMD}& 3D Vision and Touch
        &            * &            * &        0.052 &        0.175 &        0.081 &        0.090 \\ 
        \multirow{4}{*}{ }  & Ours (Vision only)
        &\textbf{0.054}&        0.028 &\textbf{0.026}&        0.083 &        0.110 &        0.051 \\ 
        \multirow{4}{*}{ }  & Ours (VTacO)
        &        0.059 &\textbf{0.018}&        0.028 &\textbf{0.012}&\textbf{0.052}&\textbf{0.028}\\
        \multirow{4}{*}{ } & Ours (VTacOH) & 0.056 & 0.024 & 0.027& 0.035 & 0.068 & 0.042\\
        \bottomrule

    \end{tabular}
    \caption{Quantitative results on YCB and AKB Objects. This table illustrates a numerical comparison among 3DVT, pure vision methods, and our approach VTacO. We measure IoU (The larger the better), Chamfer Distance ($\times 100$, the smaller the better), and Earth Mover's Distance (The smaller the better). CD and EMD are measured at 2048 points. * indicates that 3DVT has no clear settings for deformable objects: Bottle and Box.}
    \label{tab:metric}
\end{table*}

\subsection{Ablation Study}\label{sec:ablative}
\begin{figure*}[t!]
    \centering
   \includegraphics[width=1\linewidth]{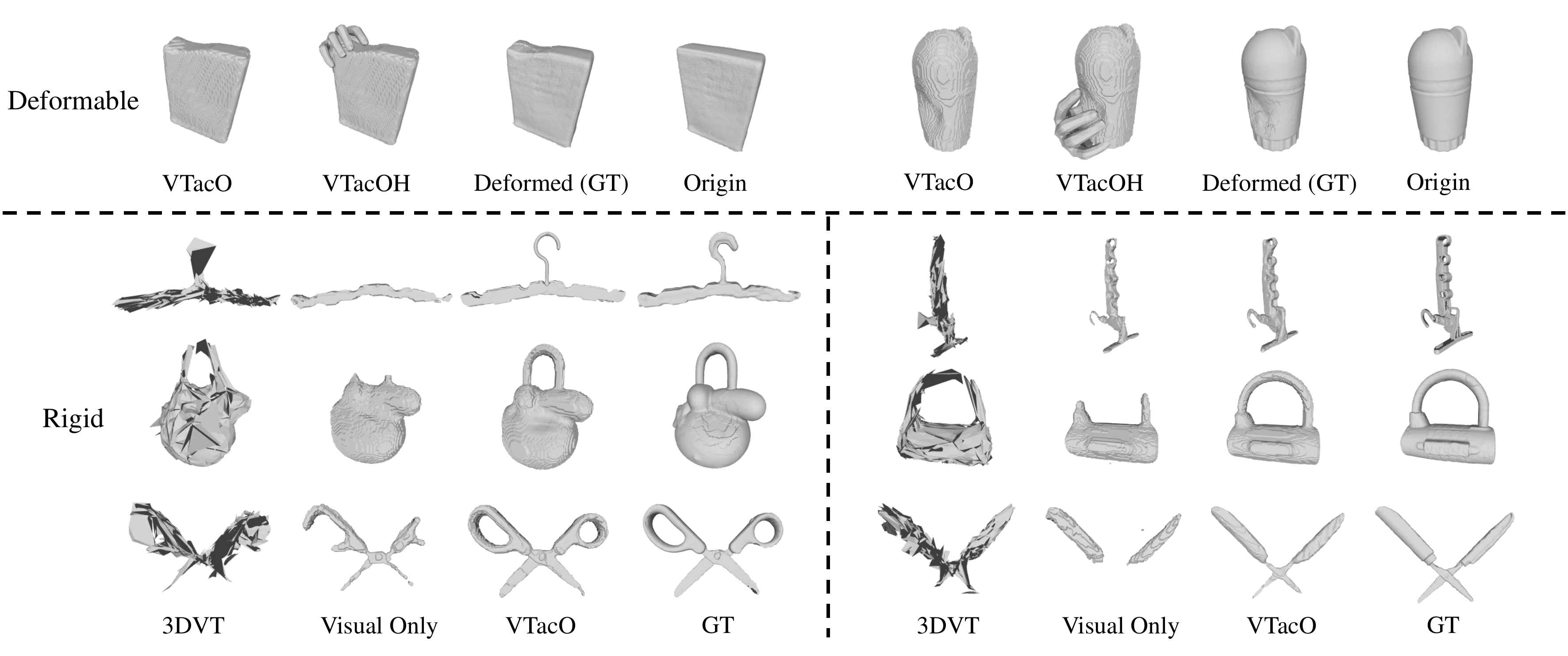}
    \caption{Qualitative example on \textbf{3DVT}, \textbf{Pure vision method}, \textbf{VTacO} and \textbf{VTacOH}. The upper demonstration shows our good performance on deformable objects. The results lower compare the visualization with different methods. We can clearly see that our method has a better qualitative reconstruction effect. In comparison with pure vision-based approaches and VTacO, we prove that the introduction of tactile signals improves the results significantly.}
    \label{fig:qual}
\end{figure*}

We conduct an ablation study on the next four aspects. We first compare the results between pure vision and the fusion of vision and tactile signals. By grasping the same object several times, we can procedurally obtain more refined structures without retraining. Finally, we study the influence brought by different choices of visual feature encoder and the feature fusion module.
\textbf{Pure vision v.s. Visual-tactile.} We can see results in Fig. \ref{fig:qual} and Tab. \ref{tab:metric} that, with the introduction of tactile images, the overall performance of reconstruction is better. Particularly, for the deformable objects \textbf{bottles} and \textbf{boxes}, the local feature can complete the information occluded by the hand, and better demonstrate the deformation of the objects. For rigid objects, touches can collect features of subtle texture from the surface of small, thin objects. For instance, the category \textbf{Foldingrack} has been reconstructed much better since the hollow structure has been discovered through DIGIT sensors, and the category \textbf{Lock} also has better performances in the upper area, after we touch and detect its upper mesh. 


\textbf{Procedural Tactile Embedding.} Our method allows us to procedurally gather the tactile embeddings and predict better reconstruction results without retraining the network. As shown in Fig. \ref{fig:teaser}, after gradually touching the object several times, the reconstruction result becomes better and better. For the rigid objects, the textures and structures become clearer when conducting more grasp, such as the gap structure of the folding rack. For deformable objects such as \textbf{Bottles}, the object geometry will be changed due to the deformation caused by the contact. Thanks to the contact geometry prediction, we can first recover the deformed geometry to its original shape and then reasons the structures to be completed. To demonstrate the deformable regions are indeed reconstructed, we also illustrate the unrecovered results in supplementary materials. Such property can also be used for reconstructing plastic object with sustainable deformation.


\textbf{Visual Feature Encoder Choices.}
We mainly compare two types of Visual Feature Encoder: \textbf{Multi-Plane Encoder} and \textbf{Volume Encoder}. \textbf{Multi-Plane Encoder} mainly project the input point cloud onto three orthogonal planes (i.e., plane $xy$, $xz$, or $yz$), and by using 2D UNet to encode the plane features and adding three features together, we obtain the multi-plane feature $\mathcal{F}_p^{plane} \in \mathbb{R}^{d_p}$ as our global feature for sample positions.

The quantitative result in Tab. \ref{tab:encoder} shows that the Volume Encoder outperforms in all metrics of all categories. This is mainly because the global feature is considered 3D information and fused with the local feature point-wisely, and the voxelization process fits the prediction of WNF, which indicates the position of each voxel relative to the surface of the object.




\begin{table}[!ht]

    \center
    \begin{tabular}{c|c|c|c|c}
        \toprule
        Metrics & Encoder & bottle & box & foldingrack \\ \hline
        \multirow{2}{*}{IoU}& Multi-Plane
        &        0.851 &        0.945 &        0.767\\ 
        \multirow{2}{*}{ }  & Volume
        &\textbf{0.887}&\textbf{0.950}&\textbf{0.782}\\
        \midrule
        
        \multirow{2}{*}{CD}& Multi-Plane
        &        1.296 &        0.732 &\textbf{0.940}\\ 
        \multirow{2}{*}{ } & Volume
        &\textbf{0.936}&\textbf{0.305}&        1.360 \\ 
        \midrule
        
        \multirow{2}{*}{EMD}& Multi-Plane
        &\textbf{0.052}&        0.030 &        0.029 \\ 
        \multirow{2}{*}{ }  & Volume
        &        0.059 &\textbf{0.018}&\textbf{0.028}\\
        \bottomrule

    \end{tabular}
    \caption{Measured metrics with \textbf{Multi-plane} and \textbf{Volume encoder}. The Volume encoder performs better in most categories.}
    \label{tab:encoder}

\end{table}

\textbf{Fusion Strategies.} We conduct experiments on two simple kinds of global and local feature fusion strategies: \textbf{concatenation} and \textbf{addition}. With concatenation, the final fused feature dimension $d=3+64$, while $d=3+32$ for the addition operation. Results in Tab. \ref{tab:Fuse} show that the method $\textbf{addition}$ has a much better performance compared to the others. We speculate that since we mark all-0 local feature vectors for the sample positions that are not near the sensor, the addition can correctly combine the two kinds of feature, while the fusion feature after the concatenation process would contain too many zeros, making it hard to predict the WNF for the decoders. 

\begin{table}[!ht]

    \center
    \begin{tabular}{c|c|c}
        \toprule
        Metrics & Fusion & bottle \\ \hline
        \multirow{2}{*}{IoU}& Addition
        &\textbf{0.9518}\\ 
        \multirow{2}{*}{ }  & Concat
        &        0.9339 \\
        \midrule
        
        \multirow{2}{*}{CD}& Addition
        &        0.305 \\ 
        \multirow{2}{*}{ } & Concat
        &\textbf{0.114}\\ 
        \midrule
        
        \multirow{2}{*}{EMD}& Addition
        &\textbf{0.018}\\ 
        \multirow{2}{*}{ }  & Concat
        &        0.028 \\
        \bottomrule

    \end{tabular}
    \caption{Quantitative results with different fusion strategies.}
    \label{tab:Fuse}

\end{table}

\subsection{Real-world Experiment}
\begin{figure}[h!]
    \centering
    \includegraphics[width=1\linewidth]{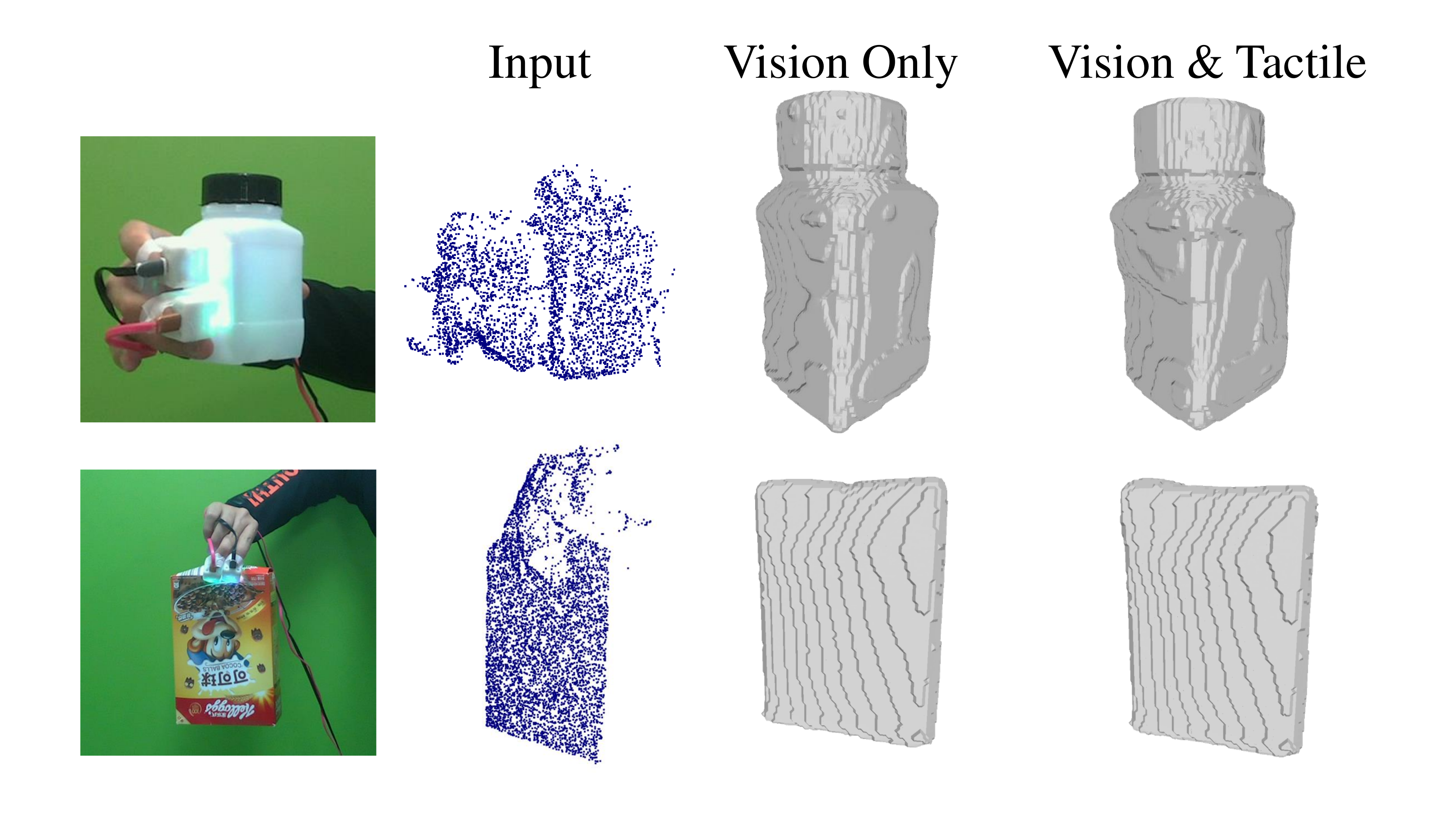}
    \caption{Real world examples. The deformation of the box and bottle has been better reconstructed with the introduction of tactile sensors in comparison with the pure-vision strategy.}
    \label{fig:real_bottle}
    \vspace{-0.3cm}
\end{figure}

We validate the performance of learned VTacO in the real-world data. We select 3 unseen \textbf{Box}, and 3 unseen \textbf{Bottle} to demonstrate the deformable object reconstruction. We also select 3 unseen \textbf{Lock} categories to demonstrate the procedural rigid reconstruction on real-world data. To note, the selected objects are not used in training from simulation. We adopt an Intel realsense L515 camera mounted on a fixed tripod to capture the depth observation. In the real world, we attach two DIGIT sensors to the hand and detect the sensor poses according to markers attached to the tips. Since we observe noise and shadows in real sensor readings, we augment the tactile images in simulation to match real RGB distribution by adding noise and adjusting contrast. We filter out the point cloud around the in-hand object manually and use it as input. In Fig. \ref{fig:real_bottle}, we illustrate examples for \textbf{Box} and \textbf{Bottle} category, as for the \textbf{Lock} category, we demonstrate them in supplementary materials.

\section{Conclusion and Future Works}
In this work, we propose a novel visual-tactile in-hand object reconstruction framework, VTacO, and its extended version, VTacOH. Our proposed frameworks can reconstruct both rigid and non-rigid objects. The visual-tactile fusion strategy allows the framework to reconstruct the object geometry details incrementally. In the future, we are interested to leverage such features in robotics experiments to better estimate the object geometry in manipulation.

\section*{Acknowledgement}
This work was supported by the National Key R\&D Program of China (No.2021ZD0110704), Shanghai Municipal Science and Technology Major Project (2021SHZDZX0102), Shanghai Qi Zhi Institute, and Shanghai Science and Technology Commission (21511101200).

{\small
\bibliographystyle{ieee_fullname}
\bibliography{vistac}
}

\end{document}